\ifdefined\pdfminorversion
\fi
\documentclass[sigconf,nonacm]{acmart}

\setcopyright{none}

\usepackage{algorithm}
\usepackage{algpseudocode}
\usepackage{array}
\usepackage{colortbl}
\usepackage{multirow}
\setkeys{acmart.cls}{balance=false}
\graphicspath{{figures/}}

\begin{document}

\title{ValQID: Learning the Net Value of Semantic Query Groups for Web Device Identification}

\author{Yifan Yang}
\affiliation{\institution{National University of Defense Technology}\city{Changsha}\country{China}}
\author{Zhen Huang}
\affiliation{\institution{National University of Defense Technology}\city{Changsha}\country{China}}
\author{Weijia Shi}
\affiliation{\institution{National University of Defense Technology}\city{Changsha}\country{China}}
\author{Xichen Tan}
\affiliation{\institution{National University of Defense Technology}\city{Changsha}\country{China}}
\author{Zhihuang Liu}
\affiliation{\institution{Xiamen University}\city{Xiamen}\country{China}}
\author{Xionglve Li}
\affiliation{\institution{National University of Defense Technology}\city{Changsha}\country{China}}
\author{Zhiping Cai}
\affiliation{\institution{National University of Defense Technology}\city{Changsha}\country{China}}
\renewcommand{\shortauthors}{Yang et al.}

\begin{abstract}
Web interfaces are widely used to access network services and manage networked devices. Identifying the vendors and types of devices behind Web interfaces is a critical step in asset inventory, security management, and risk assessment. Device identity cues are dispersed across page text, configuration fields, and management functions, and interpreting them often requires linking related Web objects. Responses and objects require different amounts of semantic analysis, so exhaustively processing all candidate content can introduce unnecessary computation. We present ValQID, which uses a relation graph to combine dispersed device identity cues, evaluates semantic questions efficiently with DSEJEV, and accounts for complementarity among semantic questions when estimating computation value. Based on the current identification state, ValQID weighs each question group's expected identification benefit against its execution cost and adaptively allocates computation to identify device vendors and types. Evaluation on real Web response data shows that ValQID achieves relative gains of up to \textbf{47.1\%} in joint macro-F1 and up to \textbf{82.1\%} in macro-F1 for low-frequency vendors over a state-of-the-art method. DSEJEV speeds up semantic processing by up to \textbf{8.58$\times$} compared with general-purpose language models.
\end{abstract}

\ccsdesc[500]{Information systems~Web mining}
\ccsdesc[500]{Networks~Network management}
\ccsdesc[100]{Security and privacy~Network security}
\ccsdesc[300]{Computing methodologies~Natural language processing}
\ccsdesc[300]{Computing methodologies~Supervised learning}

\keywords{Web device identification, Network measurement, Relational semantic analysis, Computation value estimation, Efficient semantic inference}

\maketitle

\section{Introduction}
\label{sec:introduction}

Web interfaces are widely used to access network services and manage networked devices~\cite{wei2025webagentr1,steiner2026webinterfaces,qi2025webrl,wu2025webdancer}. Networked devices such as routers, gateways, cameras, and storage systems commonly provide management functions through these interfaces~\cite{yang2026webgeoinfer,liu2025eagleye,guo2026falconscope}. Identifying the vendors and types of these devices is fundamental to network asset visibility and security risk assessment~\cite{durumeric2025censys,chen2025versionseek,sasaki2025infected}. Product identifiers, service information, and functional descriptions in device Web responses provide cues for inferring device attributes~\cite{lei2025deepfw,song2026iotfine}. Automated identification from these cues can turn large collections of Web responses into structured network asset information~\cite{durumeric2025censys,tian2025blmprobe,nasr2023chargeprint}.

\begin{figure}[!ht]
  \centering
  \includegraphics[width=\columnwidth]{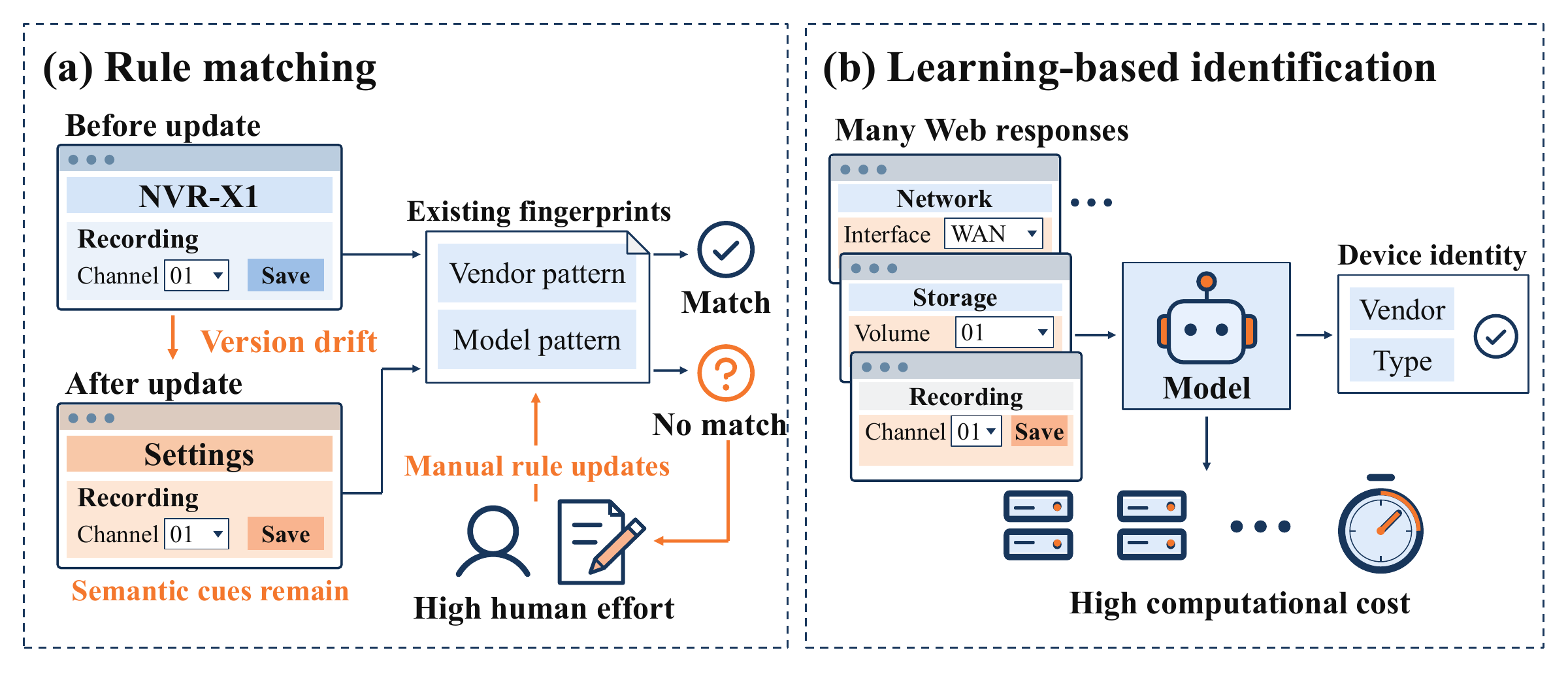}
  \caption{\textbf{Motivation for ValQID.} (a) Interface updates can invalidate fingerprints while semantic cues remain, requiring manual rule maintenance. (b) Learning-based identification uses semantic cues, with inference costs accumulating across Web responses.}
  \label{fig:motivation}
  \Description{Two panels illustrate the motivation for ValQID. In panel (a), a device interface update replaces a product identifier with a generic title while recording controls remain. Existing fingerprints no longer match, requiring manual rule updates. In panel (b), a learning-based model identifies device vendors and types from Web responses, with computing resources and a timer indicating the cost of processing many responses.}
\end{figure}

\begin{table*}[t]
  \caption{Comparison of device identification methods. $\checkmark$ and $\times$ indicate feature presence and absence. Adaptation denotes fine-tuning; Relations covers entity, structural, or cross-response links. Cost-aware selection weighs expected gains against execution costs at runtime.}
  \label{tab:related-work-comparison}
  \centering
  \begingroup
  \fontsize{9}{11}\selectfont
  \setlength{\tabcolsep}{\dimexpr0.065\textwidth/12\relax}
  \renewcommand{\arraystretch}{1.08}
  \begin{tabular*}{\textwidth}{@{\extracolsep{\fill}}
    >{\raggedright\arraybackslash}m{0.16\textwidth}
    >{\raggedright\arraybackslash}m{0.31\textwidth}
    >{\centering\arraybackslash}m{0.131\textwidth}
    >{\centering\arraybackslash}m{0.11\textwidth}
    >{\centering\arraybackslash}m{0.07\textwidth}
    >{\centering\arraybackslash}m{0.068\textwidth}
    >{\centering\arraybackslash}m{0.086\textwidth}@{}}
    \toprule
    \multirow{2}{*}[-2pt]{Method} & \multirow{2}{*}[-2pt]{Identification technique} &
    \multicolumn{3}{c}{Identification modeling} &
    \multicolumn{2}{c}{Computation control} \\
    \cmidrule(lr){3-5}\cmidrule(l){6-7}
    & & Language models & Adaptation & Relations & Adaptive & Cost-aware \\
    \midrule
    ARE~\cite{feng2018are} & Association rule mining &
    $\times$ & $\times$ & $\checkmark$ & $\times$ & $\times$ \\
    \addlinespace[1.5pt]
    WYSIWYG~\cite{wang2022wysiwyg} & CNN classification; OCR text matching &
    $\times$ & $\times$ & $\times$ & $\times$ & $\times$ \\
    \addlinespace[1.5pt]
    DevTag~\cite{wan2023devtag} & Entity and relation extraction &
    $\times$ & $\times$ & $\checkmark$ & $\times$ & $\times$ \\
    \addlinespace[1.5pt]
    LDF~\cite{sarabi2023llmfingerprint} & Contrastive RoBERTa; clustering &
    $\checkmark$ & $\checkmark$ & $\checkmark$ & $\times$ & $\times$ \\
    \addlinespace[1.5pt]
    BLMProbe~\cite{tian2025blmprobe} & LLM labeling; BERT classification &
    $\checkmark$ & $\checkmark$ & $\checkmark$ & $\times$ & $\times$ \\
    \midrule
    \rowcolor{black!10}
    \textbf{ValQID}~{\fontsize{7}{8}\selectfont(Ours)} & \textbf{Query groups; selection by net value} &
    $\boldsymbol{\checkmark}$ & $\boldsymbol{\checkmark}$ & $\boldsymbol{\checkmark}$ & $\boldsymbol{\checkmark}$ & $\boldsymbol{\checkmark}$ \\
    \bottomrule
  \end{tabular*}
  \endgroup
\end{table*}

Methods for identifying network devices behind Web interfaces commonly use rule matching or learning-based models. Rule-based methods such as ARE generate identification fingerprints by associating response terms with product descriptions~\cite{feng2018are}. Their coverage depends on the fingerprint database, which \emph{must be updated as new devices appear and response content changes} (Figure~\ref{fig:motivation})~\cite{feng2018are,kondracki2024smudged}. Learning-based methods extract identification features from response text or Web login screenshots, reducing reliance on explicit rules~\cite{wan2023devtag,wang2022wysiwyg}. BLMProbe uses language models to extract and verify device attributes from protocol responses~\cite{tian2025blmprobe}.

Applying semantic analysis to large-scale device identification presents two challenges:

\textbf{Challenge 1: Semantic understanding of dispersed identity cues.} When Web responses lack explicit vendor or device names, identification relies on cues dispersed across service descriptions, configuration fields, and management functions. For example, a configuration field's name alone may not reveal the device function it represents; interpreting it requires context from the associated managed object, enclosing form, and submission action. \emph{Matching terms or analyzing isolated fragments can overlook these relationships and miss information useful for device identification.}~\cite{li2022markuplm,zhang2023structor}

\textbf{Challenge 2: Computational efficiency of large-scale semantic identification.} As shown in Figure~\ref{fig:motivation}, inference cost grows with the number of responses, while the extent of analysis needed varies across responses~\cite{chen2024frugalgpt}. Explicit product identifiers can directly support identification, whereas interpreting the functions expressed by configuration fields may require context. Pages also contain information unrelated to device identity. Performing full semantic analysis on every response adds processing overhead for responses with sufficiently clear identity cues and for irrelevant fragments. \emph{This requires controlling the cost of each inference while adjusting the scope of semantic analysis to identification needs~\cite{ong2025routellm,zeighami2025bargain}.}

The non-generative decision model JEV offers a way to reduce semantic analysis costs~\cite{almeida2026jev}. It evaluates predefined questions in context and directly returns probability distributions over candidate answers, avoiding autoregressive text generation and answer parsing. Combining these judgments with a Web object relation graph and allocating computation by complementary question groups' expected identification benefit and execution cost offers a viable approach to device identification.

We propose ValQID, which combines relational semantic analysis with adaptive computation to identify device vendors and types from Web responses. To interpret dispersed identity cues, ValQID represents Web objects and their connections in a relation graph, using these relations to \textbf{provide local context for semantic questions and organize complementary question groups}. To improve efficiency at scale, it uses the non-generative semantic evaluator DSEJEV to reduce the cost of individual judgments and \textbf{allocates computation according to each group's expected identification benefit and execution cost}, given the current identification state. ValQID adds the judgments to the relation graph and combines the updated graph representation with character features from the response to predict device identity labels.

Our main contributions are as follows:
\begin{enumerate}
  \item \textbf{We propose ValQID, a device identification method based on computation value.} It incorporates complementarity among semantic questions into value estimation, reducing semantic queries by up to \textbf{53.16\%} while maintaining identification performance.
  \item \textbf{We design a relation graph that integrates Web object relations and semantic judgments.} Predictions from this graph are combined with character-based scores for joint device identification.
  \item \textbf{We develop DSEJEV, an efficient semantic evaluator for device identification.} We adapt a non-generative decision model to device-related semantic questions and achieve speedups of up to \textbf{8.58$\times$} in semantic processing.
  \item \textbf{We achieve state-of-the-art device identification performance.} ValQID performs best on all evaluated identification metrics across both datasets. Compared with a state-of-the-art method, it achieves relative gains of up to \textbf{47.1\%} in joint macro-F1 and up to \textbf{82.1\%} in macro-F1 for low-frequency vendors.
\end{enumerate}

\section{Related Work}
\label{sec:related-work}

Web device identification infers device identity from protocol responses and management interfaces~\cite{wan2023devtag,zhang2026nsfirmid}. Related research examines how to represent this evidence~\cite{lei2025deepfw,song2026iotfine}, select relevant content, and control the cost of acquiring and processing it~\cite{durumeric2025censys,chen2025versionseek}. Table~\ref{tab:related-work-comparison} compares representative device identification methods along two dimensions: identification modeling and computation control.

\subsection{Web Device Identification}
\label{sec:related-work-device}

Content-based methods derive fingerprints from textual or structural patterns. ARE associates response terms with product descriptions to mine identification rules~\cite{feng2018are}; fine-grained firmware fingerprinting derives DOM vectors from Web files and verifies versions through HTTP request--response comparisons~\cite{li2022securityfirmware}. Visual identification uses login-page appearance: WYSIWYG predicts vendors from screenshots and extracts device types and models through optical character recognition and text matching~\cite{wang2022wysiwyg}. Where component reuse and rebranding produce similar responses, IoTXray links devices to their original manufacturers using multiple sources~\cite{yan2023iotxray}.

Language models learn representations of response semantics and infer device attributes~\cite{zhang2026netidgpt}. LDF uses contrastively trained RoBERTa embeddings to cluster service banners and construct textual fingerprints~\cite{sarabi2023llmfingerprint}. For label construction, BLMProbe assigns extraction and verification roles to large language models, then transfers labels through protocol associations across ports~\cite{tian2025blmprobe}.

\subsection{Structural Webpage Representations and Content Selection}
\label{sec:related-work-web}

Webpage representation models preserve the context of text by encoding its location in the document hierarchy~\cite{liu2025multiui,zhang2023structor}. MarkupLM encodes node positions with XPath and learns node relationships during pretraining~\cite{li2022markuplm}. WebFormer represents relations among HTML nodes through graph attention and connects nodes to their text through node--text attention~\cite{wang2022webformer}. LCoW learns a contextualization module that extracts task-relevant interface elements and explains their functions for a decision-making agent~\cite{lee2025lcow}.

Structural information also guides the selection of webpage content for downstream tasks~\cite{kerboua2026focusagent}. In web interaction, MindAct filters candidate elements with a small model before a large model selects the target and predicts an action~\cite{deng2023mind2web}. AgentOccam simplifies webpage observations and retains pivotal elements together with their structural neighborhoods~\cite{yang2025agentoccam}. Prune4Web delegates selection to a program that traverses the DOM and applies a scoring template with model-generated, task-specific keywords and weights~\cite{zhang2026prune4web}. For retrieval-augmented question answering, HtmlRAG first filters an HTML block tree by embedding similarity, then prunes it with path generation probabilities while retaining the selected blocks' hierarchical context~\cite{tan2025htmlrag}.

\subsection{Cost-Aware Semantic Decision and Execution}
\label{sec:related-work-cost}

Adaptive feature acquisition weighs the predictive benefit of additional observations against their acquisition cost~\cite{shim2018jafa,covert2023dynamic,guney2025explainability,kobayashi2026l2m}. For individual features, DIME estimates value through conditional mutual information~\cite{gadgil2024dime}, and SEFA uses class-weighted gradient norms in a stochastic latent representation~\cite{norcliffe2025sefa}. To account for jointly informative features, ACO evaluates expected loss and cost over future feature combinations~\cite{valancius2024aco}, while Information Templates learns templates of feature combinations for instance-level acquisition~\cite{huang2025informationtemplates}.

Semantic judgments call for efficient evaluation and outputs that programs can combine directly. TypeSafe introduced Jev, a System One model for binary verification, choice, and ordinal scoring~\cite{almeida2026jev}. Its interface evaluates independent questions over shared context and returns typed probabilities for classification, scoring, and control flow~\cite{typesafe2026systemone}. LLM-as-Jev realizes this interface through the token probabilities of candidate identifiers, studying both training-free readout and candidate-selection fine-tuning in general-purpose language models~\cite{li2026llmasjev}.

Semantic data systems address cost at the level of operators and execution plans~\cite{shankar2025docetl,lin2025zendb,wei2026moar}. LOTUS expresses filtering, joining, and sorting as semantic operators and optimizes execution with proxy models, cascades, and approximation algorithms. Statistical accuracy guarantees are given at the operator level relative to reference algorithms~\cite{patel2025lotus}. Palimpzest and its Abacus optimizer select physical plans based on output quality, latency, and monetary cost~\cite{liu2025palimpzest,russo2026abacus}. Sema reduces inference cost through expression compression, operator fusion, prompt batching, and adaptive query execution~\cite{qi2026sema}. Related studies protect LLM inputs through risk-aware privacy preservation~\cite{liu2026riskaware} and semantic-aware text sanitization~\cite{zhang2025dyntext}.

\section{Methodology}
\label{sec:method}

\begin{figure*}[t]
  \centering
  \includegraphics[width=\textwidth]{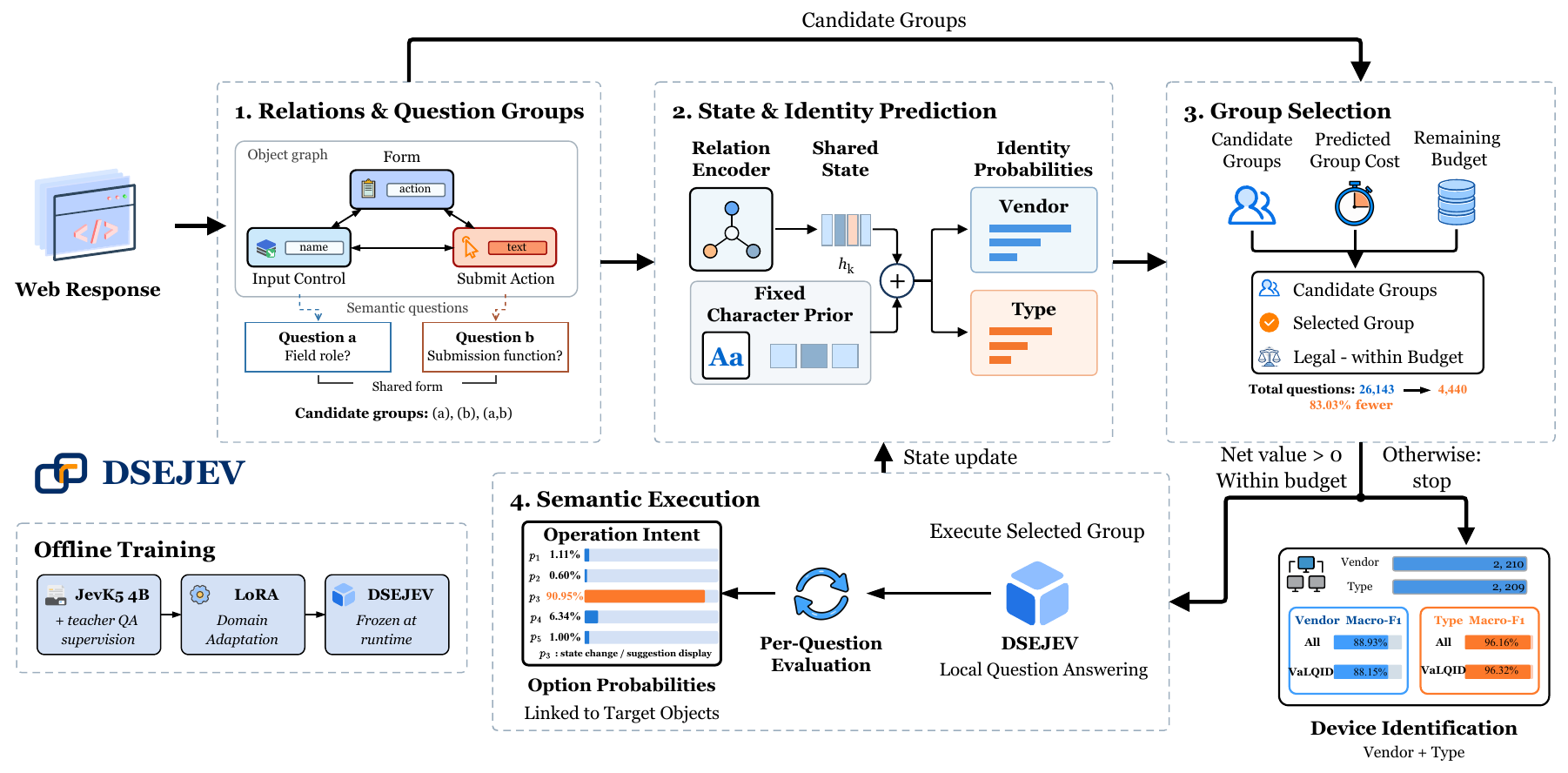}
  \caption{\textbf{ValQID workflow.} Web object relations organize complementary semantic question groups. A shared relation encoder supports identity prediction and group value estimation. ValQID selects groups by predicted net value and updates the graph with DSEJEV's option probabilities.}
  \label{fig:workflow}
  \Description{Web object relations define semantic question groups and provide the initial graph for the encoder. The encoded state and fixed character-branch scores support vendor and device type prediction. Group selection uses shared state and candidate-group representations, predicted execution cost, remaining question budget and rounds, and the cost weight. DSEJEV executes the selected questions and returns option probabilities through the state update connection to the graph encoder. Stopping returns the current vendor and device type predictions. The offline training panel shows adaptation of the semantic evaluator, which remains frozen during identification.}
\end{figure*}

ValQID uses relations among Web objects to organize semantic question groups, then decides whether to execute a group or stop based on each group's expected identification benefit and execution cost. Figure~\ref{fig:workflow} outlines the workflow.

\subsection{Problem Formulation and Overview}
\label{sec:method-overview}

Given a Web observation $x$ extracted from a target device's HTTP response, we predict its vendor and device type. We denote the true labels by $y=(y_v,y_d)$ and the predictions by $\hat y=(\hat y_v,\hat y_d)$. Interpreting a Web field often requires knowing the role of the object it belongs to and its associated operations. Evaluating semantic questions adds computational cost; the benefit to identity prediction depends on the field contents and the results already obtained. We study cost-sensitive sequential decision making over a fixed observation $x$: which groups of semantic questions to execute and when to stop.

ValQID uses Web object relations to provide local question context and organize candidate question groups. Offline screening retains semantic conditions and relation templates with individual or joint identification benefits. We learn computation value from the reduction in identity loss after adding a group's results, together with the group's execution cost, and use this value to allocate semantic computation at the current state.

Let $s_k$ denote the state before decision round $k$, containing the initial relation graph and the option probabilities obtained so far. Policy $\pi$ selects a question group $g_k$ or stops. Our semantic evaluator, DSEJEV (Device Semantic Evaluation JEV), returns option probabilities for the selected questions. Attaching these results to their target objects forms $s_{k+1}$ and updates the identity prediction. Upon stopping, joint decoding returns the labels $\hat y$.

We seek a policy $\pi$ that minimizes expected identity loss and weighted execution cost:
\begin{equation}
\mathcal J(\pi)=\mathbb E\left[\ell_y(s_T)+\lambda\sum_{k=0}^{T-1}\bar c(g_k)\right].
\label{eq:method-objective}
\end{equation}
Here $T\leq H$ is the number of executed rounds, $H$ is the maximum number of decision rounds, and $s_T$ is the terminal state. With $T=0$, the model predicts from the initial Web observation without executing semantic questions. The identity loss $\ell_y(s_T)$ is computed from the probabilities assigned to the true vendor and type labels; $\bar c(g_k)$ is the normalized group execution cost, and $\lambda\geq0$ weights that cost. The expectation is over data samples and randomness in the execution process.

\subsection{Relation-Constrained Semantic Question Groups}
\label{sec:method-groups}

ValQID parses the objects and connections in $x$ into a typed relation graph $G_x=(V_x,E_x,\tau_V,\tau_E)$. Nodes store object contents, edges record relations that can be parsed from the observation, and $\tau_V$ and $\tau_E$ specify node and edge types. Deterministic graph extraction records which object each text segment belongs to and preserves segment boundaries and order. We use these relations to assemble the local context for each semantic question.

\noindent\textbf{Question groups.} We select conditions and relation templates through offline screening on training data. A condition specifies a semantic question, a finite option set, and the objects to which it applies. Binding a condition to an object $u_a$ creates an atomic question $a=(u_a,q_a,O_a,E_a)$, where $q_a$ is the question, $O_a$ includes an insufficient-evidence option, and $E_a$ contains the target object's content and the local context needed to interpret it. For an input control, this context can include its enclosing form and the associated submission operation.

A relation template $t$ specifies the object roles associated with two questions, their connections, and the order of those roles. Two distinct atomic questions form a relation-constrained question group when their target objects match a template. The targets may be the same object, be connected by a typed path, or share a form, operation, or resource. For example, questions about a control's field role and a submission operation's function form a relation-constrained group when the two objects share a form. Templates match roles and relations and can therefore be instantiated across different page layouts.

\noindent\textbf{Condition discovery.} Using five-fold cross-validation grouped by content, we identify confusable class pairs from the initial identity predictor's out-of-fold errors. A generative model examines the misclassified training examples and their object relations to propose candidate conditions and templates. We bind these conditions to eligible objects in training observations and use the base JevK5 model to obtain their option probabilities.

Logistic regression probes assess whether these probabilities improve identity prediction, using five-fold cross-validation grouped by content. The base features contain the out-of-fold initial identity prediction, candidate identifier, and option-validity masks. Each candidate contributes one applicable binding per observation; for a relation template and its condition pair, this binding yields questions $a$ and $b$. Using the same bindings and folds, we compare four configurations: the base features alone, with the option probabilities of $a$, with those of $b$, and with both probability vectors and their outer product. The outer product represents interactions between the questions. Averaging the out-of-fold identity losses over the same applicable observations yields $L_0$, $L_a$, $L_b$, and $L_{ab}$, respectively.

We retain an individual condition when adding its option probabilities reduces identity loss. A relation template is retained when its matched question pairs lower the average loss relative to the base features, $L_{ab}<L_0$, and each question alone, $L_{ab}<\min(L_a,L_b)$, with positive interaction gain:
\begin{equation}
\Gamma(a,b)=L_a+L_b-L_{ab}-L_0>0.
\label{eq:method-interaction}
\end{equation}
This gain measures how much the joint loss reduction exceeds the sum of the two individual reductions. These criteria select templates with complementary identification evidence for the candidate catalog. We also retain their required conditions and use development data to select the library size. The retained conditions specify which local judgments the evaluator learns (Section~\ref{sec:method-learning}).

For the current observation $x$, we bind selected conditions to eligible objects and organize the resulting questions using selected relation templates to form the candidate catalog $\mathcal G_x$. The catalog contains singletons $(a)$ and relation-constrained groups $(a,b,t)$, with $a\neq b$. The catalog and each atomic question's local request remain fixed throughout identification.

\subsection{Semantic Execution, State Update, and Identity Prediction}
\label{sec:method-evidence}

During identification, DSEJEV executes the local questions in the selected group. We obtain DSEJEV by adapting the 4B JevK5 decision model~\cite{jevk52026model} on local questions instantiated from the selected conditions (Section~\ref{sec:method-learning}). Given the local input $(E_a,q_a,O_a)$ of atomic question $a$, DSEJEV computes option logits $\mathbf z_a$ and returns
\[
\mathbf p_a=\operatorname{softmax}_{O_a}(\mathbf z_a/\tau),
\]
where $\tau$ is a calibration temperature. The two atomic questions in a relation-constrained group are evaluated separately using their respective fixed local requests. We add both complete option probability distributions in a single state update.

A text encoder using frozen JevK5 input embeddings encodes each object's content. Each completed question adds a result node encoding its full option-probability vector and linked to its target object. State $s_k$ records the fixed observation $x$, initial graph $G_x$, completed results $Z_k$, action history, remaining question budget $b_k$, and remaining rounds $r_k$. Let $G_x\oplus Z_k$ denote the graph with the result nodes added. The relation encoder aggregates object and result representations along typed edges, then combines the graph-level representation with the budget and remaining rounds:

\begin{equation}
(h_k,\{h_{u,k}\})=F_\theta(G_x\oplus Z_k,b_k,r_k).
\label{eq:method-state}
\end{equation}
Here $h_k$ is the shared graph-level state representation, and $h_{u,k}$ represents object $u$ in the current state. These object representations also enter group value estimation.

The relation branch predicts from $h_k$ and updates its output as semantic results arrive. The character branch computes fixed classification scores from character features of the full HTTP response once per observation. We denote the task-level distributions used for identity-loss computation by $p_v(\cdot\mid s_k)$ and $p_d(\cdot\mid s_k)$. In configurations with task-level fusion, weighted character log probabilities are added to the relation logits before softmax to obtain these distributions.

\begin{samepage}
During offline training, we use the training labels to compute the identity loss from these probabilities:
\[
\ell_y(s)=m_v\ell_v(s)+\omega_d m_d\ell_d(s).
\]
For task $j\in\{v,d\}$, $\ell_j(s)=-\log p_j(y_j\mid s)$. The masks $m_v,m_d\in\{0,1\}$ indicate valid labels, and $\omega_d$ weights the device type task. The decrease in this loss after executing a question group measures the identification benefit of its semantic results. Together with execution cost, it provides supervision for group value learning in Section~\ref{sec:method-selection}.
\end{samepage}

When the policy stops at $s_T$, joint decoding combines the task scores and fixed joint evidence:
\[
S(y\mid x,s_T)=A_v(y_v\mid x,s_T)+A_d(y_d\mid x,s_T)+C(y\mid x).
\]
Here $A_v$ and $A_d$ are the terminal vendor and type scores after character fusion and scaling, carrying forward any character contributions already present in $p_v,p_d$. The fixed term $C(y\mid x)$ combines scores from character-based linear classifiers trained on vendor--type pairs with label-pair associations and frequency corrections from the training data. In the DevTag configuration, it also incorporates labels from similar training observations. The task-level loss supervises computation selection; joint decoding produces the final labels:
\[
\hat y=\arg\max_y S(y\mid x,s_T).
\]

\subsection{Group Value and Computation Selection}
\label{sec:method-selection}

Each observation has a maximum of $H$ decision rounds and a budget of $B$ atomic questions. Each round executes one question group and updates the state. The legal group set $\mathcal G(s)\subseteq\mathcal G_x$ contains groups whose questions have not been executed and fit the remaining question budget. When no rounds remain, $\mathcal G(s)=\varnothing$.

A value model estimates each legal group's net value at the current state to select a group or stop.

\noindent\textbf{Group execution cost.} We measure a group's amortized execution time $c(g)$ under fixed reference batching and inference settings and normalize it by a reference cost $c_0>0$: $\bar c(g)=c(g)/c_0$. We train a group cost predictor on measured execution costs of singletons and relation-constrained groups. Value training targets use measured normalized costs $\bar c(g)$. The cost predictor supplies the normalized estimate $\widehat c(g)/c_0$ to the value head before group selection.

\noindent\textbf{Value learning.} Let $s^g$ be the state after executing $g$ and attaching all its option probabilities to their objects. The loss reduction from $s$ to $s^g$ evaluates the group's results together, allowing complementary questions to contribute jointly to its value. Subtracting the weighted execution cost gives the immediate net return:
\begin{equation}
R_\lambda(s,g)=\ell_y(s)-\ell_y(s^g)-\lambda\bar c(g).
\label{eq:method-reward}
\end{equation}
With one round remaining, the value target is $Y_1(s,g)=R_\lambda(s,g)$. Let $\mathcal S(s^g)\subseteq\mathcal G(s^g)$ contain legal successor groups with recorded semantic outputs, execution costs, and state transitions. For $1<r\leq H$, we construct the target recursively from these transitions:
\begin{equation}
\begin{aligned}
Y_r(s,g)={}&R_\lambda(s,g)\\
&+\max\left(\{0\}\cup\{Y_{r-1}(s^g,g'):g'\in\mathcal S(s^g)\}\right).
\end{aligned}
\label{eq:method-value-targets}
\end{equation}
The zero-valued option corresponds to stopping; an empty successor set gives zero continuation return.

Summing Equation~\eqref{eq:method-reward} along a trajectory yields
\[
\sum_{k=0}^{T-1}R_\lambda(s_k,g_k)=\ell_y(s_0)-\ell_y(s_T)-\lambda\sum_{k=0}^{T-1}\bar c(g_k).
\]
Once the parameters and task-level fusion weights used to compute $\ell_y$ are fixed, $\ell_y(s_0)$ is independent of the policy. Maximizing the expected cumulative net return therefore corresponds to minimizing Equation~\eqref{eq:method-objective}.

The shared value head $Q_\phi$ uses $h_k$, target-object and condition encodings, and, for relation-constrained groups, the template encoding and interaction features between the two questions. It also receives the remaining question budget, a round condition, normalized cost, and cost weight $\lambda$. Trained against $Y_r$, it predicts group $g$'s net value $\widehat Q_k(g)$ at round $k$. During identification, one-step evaluation sets the round input to $1$ at every decision; evaluation based on remaining rounds uses $r_k$.

If $\mathcal G(s_k)$ is empty, the policy returns the current identity prediction. Otherwise, it selects the group with the highest predicted net value:
\begin{equation}
g_k^\star=\arg\max_{g\in\mathcal G(s_k)}\widehat Q_k(g).
\label{eq:method-decision}
\end{equation}
The group is executed when $\widehat Q_k(g_k^\star)>0$; otherwise, the model returns its current identity prediction. After execution, the results update the state, the question budget decreases by the number of completed questions, and the remaining rounds decrease by one before the next decision.

Algorithm~\ref{alg:online-identification} summarizes the identification procedure.

\begin{algorithm}[t]
\caption{ValQID Identification}
\label{alg:online-identification}
\begin{algorithmic}[1]
\Require Web observation $x$, condition and template library, trained models, maximum decision rounds $H$, question budget $B$, cost weight $\lambda$, value-evaluation mode
\Ensure Identity labels $\hat y=(\hat y_v,\hat y_d)$
\State Build $G_x$, $\mathcal G_x$, and fixed local requests.
\State Compute and retain character-branch outputs and joint-score contributions that depend only on $x$.
\State Initialize $s_0$ with $Z_0=\varnothing$, $b_0=B$, $r_0=H$, and an empty action history; $k\gets0$.
\Loop
  \State $(h_k,\{h_{u,k}\})\gets F_\theta(G_x\oplus Z_k,b_k,r_k)$.
  \State Compute the current vendor and type distributions $p_v(\cdot\mid s_k)$ and $p_d(\cdot\mid s_k)$.
  \State Form $\mathcal G(s_k)$ from groups with no completed questions and at most $b_k$ questions.
  \State Set $\mathcal G(s_k)\gets\varnothing$ if $r_k=0$.
  \State \textbf{if} $\mathcal G(s_k)=\varnothing$ \textbf{then} \Return $\hat y=\arg\max_y S(y\mid x,s_k)$.
  \State Predict $\widehat c(g)/c_0$ for each $g\in\mathcal G(s_k)$.
  \State Use $Q_\phi$ to obtain each $\widehat Q_k(g)$ from shared representations, group features, $b_k,\widehat c(g)/c_0,\lambda$, with round input $1$ for one-step evaluation and $r_k$ otherwise.
  \State $g_k^\star\gets\arg\max_{g\in\mathcal G(s_k)}\widehat Q_k(g)$.
  \State \textbf{if} $\widehat Q_k(g_k^\star)\leq0$ \textbf{then} \Return $\hat y=\arg\max_y S(y\mid x,s_k)$.
  \For{each atomic question $a$ in $g_k^\star$}
    \State $\mathbf p_a\gets\operatorname{DSEJEV}(E_a,q_a,O_a)$.
  \EndFor
  \State Form $Z_{k+1}$ by adding all returned $\mathbf p_a$ to $Z_k$, each attached to its target object $u_a$.
  \State Record the completed questions in the action history.
  \State $b_{k+1}\gets b_k-\text{number of questions in }g_k^\star$.
  \State $r_{k+1}\gets r_k-1$; form $s_{k+1}$; $k\gets k+1$.
\EndLoop
\end{algorithmic}
\end{algorithm}

\subsection{Training and Identification}
\label{sec:method-learning}

We fix the condition and template library selected in Section~\ref{sec:method-groups} and bind the retained conditions to eligible objects in the training observations. Each adaptation example contains a question $q_a$, finite options $O_a$, and local context $E_a$, which includes the target object's content and the relations needed to interpret it. We use Qwen3.5-397B-A17B as the teacher to select an answer from $O_a$ based on the local context $E_a$, providing evidence references and a brief rationale. Three researchers manually review all annotations to check whether the selected answers are supported by the local evidence. We fine-tune the base JevK5 model on these labeled examples using LoRA~\cite{hu2022lora} to obtain DSEJEV.

We then freeze DSEJEV and use its option probabilities to train the relation encoder and identity predictor. Training includes initial states and states augmented with singleton or relation-constrained group results, whether obtained in one round or accumulated across rounds.

After training the identity predictor, we select task-level fusion weights on development data, then freeze the relation encoder, task heads, and these weights. On a separate value-fitting subset, we compute task-level loss reductions and measured execution costs for recorded transitions to construct the corresponding targets $Y_r$. These targets train the candidate encoder and value head. For each data source, we follow the same training procedure with a separate label space, condition library, and set of model parameters. Deployment settings are selected on development data.

We then fit the joint linear classifiers using training samples with valid vendor and type labels, keeping the previously trained models fixed. Joint-decoding weights are selected on development data and remain fixed during identification.

\noindent\textbf{Identification.} Observations independently select question groups using the policy in Section~\ref{sec:method-selection}. Selected groups are collected into batches for DSEJEV execution, and the returned probabilities are attached to their target objects in each observation. Each observation updates its state and continues or stops independently according to Algorithm~\ref{alg:online-identification}.

\raggedbottom
\section{Experiments}
\label{sec:experiments}

We evaluate ValQID's identification performance and efficiency through the following questions:

\noindent\textbf{RQ1:} How does ValQID compare with existing methods overall and on long-tail vendors?

\noindent\textbf{RQ2:} How much does DSEJEV improve semantic processing speed and overall throughput?

\noindent\textbf{RQ3:} How do identification performance and computation cost change when key components are removed?

\begin{table*}[!t]
  \definecolor{comparisonred}{HTML}{B32624}
  \newcommand{\reldrop}[1]{{\fontsize{7}{8}\selectfont\textcolor{comparisonred}{$\downarrow$\,#1\%}}}
  \newcommand{\metriccell}[2]{\makebox[18.2pt][r]{#1}\hspace{0.5pt}\makebox[20.5pt][l]{\reldrop{#2}}}
  \newcommand{\ourscell}[1]{\makebox[39.2pt][l]{\makebox[18.2pt][r]{\textbf{#1}}}}
  \newcommand{\methodcite}[1]{{\fontsize{7}{8}\selectfont\cite{#1}}}
  \caption{Identification performance on DevTag and BLMProbe (\%). Acc.\ denotes accuracy; Prec., Rec., and M-F1 denote macro-averaged precision, recall, and F1. Joint metrics evaluate vendor--type pairs. $\uparrow$ indicates higher is better. Red $\downarrow$ values are relative decreases from ValQID, $100(s_{\mathrm{ours}}-s)/s_{\mathrm{ours}}$; bold marks the best score for each metric and dataset.}
  \label{tab:main-comparison}
  \centering
  \fontsize{8.5}{11}\selectfont
  \setlength{\tabcolsep}{0.8pt}
  \renewcommand{\arraystretch}{1}
  \begin{tabular}{l>{\hspace{3pt}}l<{\hspace{3pt}}*{10}{>{\raggedleft\arraybackslash}p{39.2pt}}}
    \toprule
    \multirow{2}{*}{Dataset} & \multirow{2}{*}{Method}
      & \multicolumn{4}{c}{Vendor Identification}
      & \multicolumn{4}{c}{Device-Type Identification}
      & \multicolumn{2}{c}{Joint Identification} \\
    \cmidrule(lr){3-6}\cmidrule(lr){7-10}\cmidrule(l){11-12}
      & & \multicolumn{1}{c}{Acc.\ $\uparrow$} & \multicolumn{1}{c}{Prec.\ $\uparrow$}
      & \multicolumn{1}{c}{Rec.\ $\uparrow$} & \multicolumn{1}{c}{M-F1 $\uparrow$}
      & \multicolumn{1}{c}{Acc.\ $\uparrow$} & \multicolumn{1}{c}{Prec.\ $\uparrow$}
      & \multicolumn{1}{c}{Rec.\ $\uparrow$} & \multicolumn{1}{c}{M-F1 $\uparrow$}
      & \multicolumn{1}{c}{Acc.\ $\uparrow$} & \multicolumn{1}{c}{M-F1 $\uparrow$} \\
    \midrule
    \multirow{5}{*}{DevTag}
      & DevTag~\methodcite{wan2023devtag} & \metriccell{57.18}{40.4} & \metriccell{51.19}{37.4} & \metriccell{47.09}{35.8} & \metriccell{45.35}{39.9} & \metriccell{58.50}{39.2} & \metriccell{39.36}{58.8} & \metriccell{28.61}{68.7} & \metriccell{31.18}{66.3} & \metriccell{55.58}{41.7} & \metriccell{38.49}{49.1} \\
      & RoBERTa~\methodcite{sarabi2023llmfingerprint} & \metriccell{89.36}{6.8} & \metriccell{38.82}{52.5} & \metriccell{33.19}{54.8} & \metriccell{33.82}{55.2} & \metriccell{89.45}{7.1} & \metriccell{46.74}{51.0} & \metriccell{44.56}{51.2} & \metriccell{45.18}{51.2} & \metriccell{86.10}{9.7} & \metriccell{27.05}{64.2} \\
      & BLMProbe~\methodcite{tian2025blmprobe} & \metriccell{92.71}{3.3} & \metriccell{58.85}{28.0} & \metriccell{59.08}{19.5} & \metriccell{55.97}{25.9} & \metriccell{93.66}{2.7} & \metriccell{73.81}{22.7} & \metriccell{84.03}{8.0} & \metriccell{77.07}{16.8} & \metriccell{91.99}{3.5} & \metriccell{56.25}{25.7} \\
      & Qwen3.5~\methodcite{qwen2026qwen35} & \metriccell{70.27}{26.7} & \metriccell{31.42}{61.6} & \metriccell{28.39}{61.3} & \metriccell{26.38}{65.1} & \metriccell{35.50}{63.1} & \metriccell{22.75}{76.2} & \metriccell{10.11}{88.9} & \metriccell{10.91}{88.2} & \metriccell{32.46}{65.9} & \metriccell{13.26}{82.5} \\
    \rowcolor{black!10}
      & \textbf{ValQID}~{\fontsize{7}{8}\selectfont(Ours)}
      & \ourscell{95.91} & \ourscell{81.76} & \ourscell{73.38} & \ourscell{75.50}
      & \ourscell{96.30} & \ourscell{95.46} & \ourscell{91.33} & \ourscell{92.62}
      & \ourscell{95.30} & \ourscell{75.65} \\
    \midrule
    \multirow{5}{*}{BLMProbe}
      & DevTag~\methodcite{wan2023devtag} & \metriccell{13.08}{86.5} & \metriccell{10.46}{88.9} & \metriccell{7.96}{91.6} & \metriccell{8.41}{91.0} & \metriccell{9.32}{90.6} & \metriccell{40.32}{58.9} & \metriccell{13.19}{86.6} & \metriccell{11.86}{87.9} & \metriccell{4.39}{95.5} & \metriccell{3.09}{96.6} \\
      & RoBERTa~\methodcite{sarabi2023llmfingerprint} & \metriccell{88.58}{8.3} & \metriccell{62.15}{33.9} & \metriccell{61.78}{34.5} & \metriccell{59.34}{36.4} & \metriccell{91.52}{7.6} & \metriccell{70.01}{28.7} & \metriccell{68.52}{30.4} & \metriccell{69.11}{29.7} & \metriccell{86.15}{10.8} & \metriccell{49.55}{46.1} \\
      & BLMProbe~\methodcite{tian2025blmprobe} & \metriccell{91.56}{5.2} & \metriccell{69.84}{25.8} & \metriccell{71.33}{24.3} & \metriccell{68.29}{26.8} & \metriccell{95.57}{3.5} & \metriccell{90.61}{7.7} & \metriccell{82.97}{15.7} & \metriccell{85.06}{13.5} & \metriccell{91.44}{5.4} & \metriccell{62.47}{32.0} \\
      & Qwen3.5~\methodcite{qwen2026qwen35} & \metriccell{51.98}{46.2} & \metriccell{52.88}{43.8} & \metriccell{52.50}{44.3} & \metriccell{47.25}{49.3} & \metriccell{79.16}{20.1} & \metriccell{87.90}{10.5} & \metriccell{65.55}{33.4} & \metriccell{68.91}{29.9} & \metriccell{47.25}{51.1} & \metriccell{36.79}{60.0} \\
    \rowcolor{black!10}
      & \textbf{ValQID}~{\fontsize{7}{8}\selectfont(Ours)}
      & \ourscell{96.63} & \ourscell{94.09} & \ourscell{94.28} & \ourscell{93.28}
      & \ourscell{99.03} & \ourscell{98.22} & \ourscell{98.48} & \ourscell{98.34}
      & \ourscell{96.63} & \ourscell{91.90} \\
    \bottomrule
  \end{tabular}
  \Description{Five methods are compared on DevTag and BLMProbe. Each dataset reports four vendor metrics, four device-type metrics, and two joint identification metrics. Precision, recall, and F1 are macro-averaged. Each baseline score is followed on the same line by a red downward arrow and its percentage decrease relative to ValQID. ValQID achieves the highest value in all ten metrics on both datasets, and its rows are shaded across the entire table.}
\end{table*}

\subsection{Datasets and Experimental Setup}
\label{sec:datasets}

We evaluate device vendor and type identification using two Web response datasets, DevTag and BLMProbe.

\textbf{DevTag.} DevTag is a public benchmark for IoT device fingerprinting, with application-layer service responses labeled by device type, vendor, and product~\cite{wan2023devtag}. We use its released preprocessed response texts. We deduplicate records by exact full-text matching, remove records with identical text but conflicting labels, and select HTTP responses based on their status lines. Vendor and device-type labels are extracted from the original annotations. The resulting HTTP dataset contains 75,188 unique text records covering 80 vendor classes and 20 device types, including cameras, routers, gateways, switches, and printers. It retains low-frequency vendor and device-type classes for evaluating long-tail identification performance.

\textbf{BLMProbe.} BLMProbe releases protocol responses collected through active probing of Internet-connected devices, together with vendor, device-type, and product labels~\cite{tian2025blmprobe}. We extract non-empty HTTP/HTTPS responses and their existing vendor and device-type labels from this data. The input retains response headers and available HTML bodies, including page text, forms, and scripts. The dataset contains 11,164 responses covering 67 vendor classes and five device types: routers, gateways, digital video recorders, switches, and printers.

The training, development, and test sets contain 58,924, 7,774, and 8,490 responses for DevTag and 6,572, 2,222, and 2,370 responses for BLMProbe, respectively. For ValQID, we further divide each training set into two disjoint subsets at an approximate 4:1 ratio. The larger subset is used to train the relation-based identity model and discover semantic conditions; the smaller subset is used to fit the question-group value model while keeping the identity model parameters fixed. We use the development set to select model checkpoints, fusion parameters, and query configurations. All methods are evaluated on the same test responses within each dataset using a common label vocabulary and scoring procedure.

\subsection{Comparison with Baselines}
\label{sec:baseline-comparison}

We compare ValQID with representative device identification methods, including DevTag~\cite{wan2023devtag}, RoBERTa~\cite{sarabi2023llmfingerprint}, and BLMProbe~\cite{tian2025blmprobe}, as well as the general-purpose large language model Qwen3.5~\cite{qwen2026qwen35}. Table~\ref{tab:main-comparison} reports vendor, device-type, and joint identification results on both datasets. We use accuracy to measure overall identification performance. Since samples from major vendors account for a large share of the data, we macro-average precision, recall, and F1 to give rare and frequent classes equal weight when evaluating performance under long-tailed class distributions.

On DevTag, which has a long-tailed class distribution, ValQID improves vendor macro-F1 by 34.9\% relative to the strongest baseline, BLMProbe. Joint identification requires both the vendor and device type to be predicted correctly for the same response. ValQID achieves joint accuracies of 95.30\% on DevTag and 96.63\% on BLMProbe, with relative improvements of 34.5\% and 47.1\% in joint macro-F1 over BLMProbe, respectively. Across both datasets, ValQID achieves the best results on every reported metric for vendors, device types, and vendor--type pairs.

\begin{table*}[!t]
  \caption{Ablation of semantic queries, character-branch fusion, and value-based query selection. Identification scores are percentages; M-F1 denotes macro-F1. Semantic query and input token counts are normalized to ValQID within each dataset ($1.00\times$), with one query per semantic question. Bold marks the best value for each metric and dataset.}
  \label{tab:ablation}
  \centering
  \newcommand{\ablationhead}[1]{{\fontsize{8}{10}\selectfont#1}}
  \fontsize{9}{11}\selectfont
  \setlength{\tabcolsep}{3pt}
  \renewcommand{\arraystretch}{1.08}
  \begin{tabular}{p{42pt}p{138pt}*{4}{>{\raggedleft\arraybackslash}p{\dimexpr(\textwidth-42pt-138pt-66pt-62pt-16\tabcolsep)/4\relax}}>{\raggedleft\arraybackslash}p{66pt}>{\raggedleft\arraybackslash}p{62pt}}
    \toprule
    \multirow{2}{*}{\ablationhead{Dataset}} & \multirow{2}{*}{\ablationhead{Configuration}}
      & \ablationhead{Vendor} & \makebox[\linewidth][r]{\ablationhead{Device type}} & \multicolumn{2}{c}{\ablationhead{Joint identification}}
      & \ablationhead{Semantic queries} & \ablationhead{Input tokens} \\
    \cmidrule(lr){3-3}\cmidrule(lr){4-4}\cmidrule(lr){5-6}\cmidrule(lr){7-7}\cmidrule(lr){8-8}
      & & \ablationhead{M-F1 $\uparrow$} & \ablationhead{M-F1 $\uparrow$} & \ablationhead{Acc.\ $\uparrow$} & \ablationhead{M-F1 $\uparrow$} & \ablationhead{($\times$) $\downarrow$} & \ablationhead{($\times$) $\downarrow$} \\
    \midrule
    \rowcolor{black!10}
    \multirow{4}{*}{DevTag} & \textbf{ValQID}
      & 75.50 & \textbf{92.63} & 95.31 & 75.66 & $1.00\times$ & $1.00\times$ \\
      & w/o Semantic Queries
      & 38.55 & 62.03 & 88.21 & 39.11 & \textemdash & \textemdash \\
      & w/o Character Branch and Fusion
      & 38.65 & 60.73 & 89.32 & 35.40 & $1.00\times$ & $1.00\times$ \\
      & w/o Value-based Query Selection
      & \textbf{75.56} & \textbf{92.63} & \textbf{95.32} & \textbf{75.70} & $1.81\times$ & $1.73\times$ \\
    \midrule
    \rowcolor{black!10}
    \multirow{4}{*}{BLMProbe} & \textbf{ValQID}
      & \textbf{93.28} & \textbf{98.34} & \textbf{96.63} & \textbf{91.90} & $1.00\times$ & $1.00\times$ \\
      & w/o Semantic Queries
      & 76.56 & 91.26 & 92.07 & 71.67 & \textemdash & \textemdash \\
      & w/o Character Branch and Fusion
      & 77.15 & 90.03 & 91.99 & 73.26 & $1.00\times$ & $1.00\times$ \\
      & w/o Value-based Query Selection
      & \textbf{93.28} & \textbf{98.34} & \textbf{96.63} & \textbf{91.90} & $2.14\times$ & $2.76\times$ \\
    \bottomrule
  \end{tabular}
  \Description{Four configurations are compared on DevTag and BLMProbe using vendor and device-type macro-F1, joint accuracy, joint macro-F1, relative semantic queries, and relative input tokens. Each semantic question counts as one query; both computation measures are normalized to the full configuration within each dataset. Disabling semantic queries yields joint macro-F1 scores of 39.11 percent on DevTag and 71.67 percent on BLMProbe, with zero semantic queries and input tokens. Removing the character branch and fusion reduces all four identification scores on both datasets. Fixed-order execution produces similar joint identification scores while requiring 1.81 times and 2.14 times as many semantic queries, and 1.73 times and 2.76 times the input tokens, on DevTag and BLMProbe, respectively. Full-configuration rows are shaded.}
\end{table*}

\subsection{Long-Tail Analysis}
\label{sec:long-tail-analysis}

\begin{figure}[!t]
  \centering
  \includegraphics[width=\columnwidth]{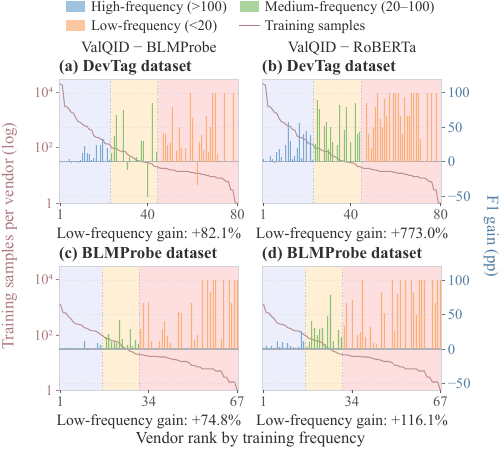}
  \caption{Vendor F1 gains by training frequency. Stems show per-vendor F1 differences in percentage points (pp). Panel labels show low-frequency macro-F1 gains as percentages of the corresponding baseline's macro-F1.}
  \label{fig:vendor-frequency}
  \Description{A single-column figure contains four panels in a two-by-two arrangement. Panels a and b are labeled DevTag dataset, and panels c and d are labeled BLMProbe dataset. The left and right columns compare ValQID with the BLMProbe and RoBERTa baselines. Vendors are ranked by decreasing training sample count. Blue, green, and orange stems show signed per-vendor F1 differences in percentage points for high-, medium-, and low-frequency groups. A thin curve shows training sample counts on the left logarithmic axis. Each panel has a one-line Low-frequency gain label reporting relative macro-F1 improvements of 82.1 and 773.0 percent on DevTag and 74.8 and 116.1 percent on BLMProbe, respectively, with vendors weighted equally.}
\end{figure}

To examine whether identification gains extend to long-tail vendors, we divide vendors in each dataset into high, medium, and low frequency groups based on their training sample counts. We compare ValQID with BLMProbe and RoBERTa using macro-F1 within each group, giving equal weight to each vendor. Figure~\ref{fig:vendor-frequency} shows per-vendor F1 gains, with vendors ordered by decreasing training sample count.

Across both datasets, ValQID's relative improvements over both baselines increase from the high frequency group to the medium and low frequency groups. For low frequency vendors, ValQID achieves macro-F1 scores of 66.28\% on DevTag and 91.81\% on BLMProbe, corresponding to relative improvements of 82.1\% and 74.8\% over the BLMProbe baseline.

Low frequency vendors account for approximately half of the vendor classes in each dataset. ValQID improves F1 over BLMProbe for 22 of 36 such vendors on DevTag and 25 of 37 on BLMProbe, covering a majority of the tail classes. ValQID combines character patterns from the full response with semantic assessments of Web objects and their relations. These inputs provide both textual and functional information for identifying vendors with limited training examples.

\subsection{Efficiency of DSEJEV}
\label{sec:executor-efficiency}

\begin{figure}[!t]
  \centering
  \includegraphics[width=\columnwidth]{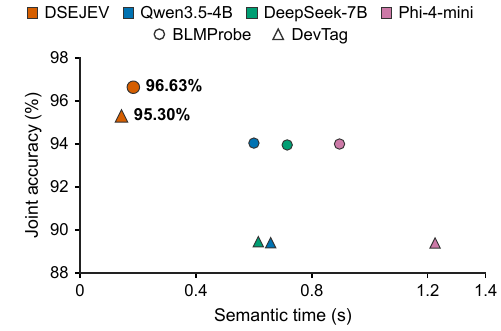}
  \caption{Joint identification accuracy (\%) versus semantic execution time (s per 1,000 responses). Circles denote BLMProbe; triangles denote DevTag.}
  \label{fig:executor-efficiency}
  \Description{Eight independent points show joint vendor and device type accuracy against semantic execution time. Colors distinguish DSEJEV, Qwen3.5-4B, DeepSeek-7B, and Phi-4-mini; circles denote BLMProbe and triangles denote DevTag. DSEJEV is labeled 96.63 percent at 0.184 seconds per 1,000 responses on BLMProbe and 95.30 percent at 0.143 seconds on DevTag. The horizontal axis spans 0 to 1.4 seconds and the vertical axis spans 88 to 98 percent.}
\end{figure}

We compare DSEJEV with three alternative semantic evaluators within ValQID: Qwen3.5-4B, DeepSeek LLM 7B Chat, and Phi-4-mini-instruct (Figure~\ref{fig:executor-efficiency}). Across both datasets, the DSEJEV configurations achieve \textbf{3.27--8.58$\times$} speedups in semantic processing and the highest joint identification accuracy among the evaluated configurations. Using DSEJEV also raises end-to-end identification throughput to \textbf{1.20--1.83} times that of the alternative evaluator configurations.

DSEJEV is built on JevK5~\cite{jevk52026model} and fine-tuned on local semantic questions for device identification. It returns probabilities over predefined options in a single forward pass, avoiding autoregressive text generation and answer parsing. These probability vectors feed directly into identity prediction and subsequent computation selection.

\subsection{Ablation Study}
\label{sec:ablation}

Table~\ref{tab:ablation} compares ValQID with three ablated configurations. For the semantic-query ablation, we disable question execution and result injection while retaining the identity model, character branch, and fusion components. Semantic queries improve joint macro-F1 by 36.55 percentage points on DevTag and 20.23 percentage points on BLMProbe.

The character-branch and fusion ablation jointly removes the character classifiers, label-pair and frequency corrections, joint classifiers, and the DevTag nearest-neighbor component. It uses the same semantic-question trace and relation-model outputs as ValQID, with vendor and device type predicted independently. Together, the character and fusion components increase joint macro-F1 from 35.40\% to 75.66\% on DevTag and from 73.26\% to 91.90\% on BLMProbe.

Compared with fixed-order execution, value-based query selection reduces semantic query counts by 44.74\% on DevTag and 53.16\% on BLMProbe, and input token usage by 42.08\% and 63.83\%, respectively, while maintaining comparable joint identification performance. Both configurations use the same question catalog, DSEJEV evaluator, identity predictor, and question budget. Fixed-order execution follows a predefined sequence until the budget is exhausted. ValQID selects the eligible question group with the highest predicted net value and stops when no eligible group has positive net value.

\section{Conclusion}
\label{sec:conclusion}

We presented ValQID, a method for identifying device vendors and types from Web responses. It accounts for complementarity among semantic questions when estimating computation value: it groups questions using Web object relations and selects groups based on their expected identification benefit and execution cost given the current identification state. On real Web response data, ValQID achieves relative gains of up to 47.1\% in joint macro-F1 and up to 82.1\% in macro-F1 for low-frequency vendors over a state-of-the-art method. DSEJEV speeds up semantic processing by up to 8.58$\times$ compared with the three language-model evaluators. These results show that assessing the value of complementary question groups allows relationships among identity cues to guide semantic computation, enabling accurate and efficient device identification. Future work could explore how to adapt semantic question groups and their value estimates as firmware and Web interface updates change identity cues.

\flushbottom
\bibliographystyle{ACM-Reference-Format}
\bibliography{references}

\clearpage
\appendix
\twocolumn[{
\begin{minipage}{\textwidth}
\begin{table}[H]
  \caption{(a) Semantic questions and relation templates. (b) Example request and DSEJEV output.}
  \label{tab:semantic-question-examples}
  \centering
  \begingroup
  \small
  \setlength{\tabcolsep}{0.01\textwidth}
  \renewcommand{\arraystretch}{1.08}
  \noindent\textbf{(a) Semantic questions and relation templates}\par\smallskip
  \begin{tabular}{@{}
    >{\raggedright\arraybackslash}p{0.035\textwidth}
    >{\raggedright\arraybackslash}p{0.185\textwidth}
    >{\raggedright\arraybackslash}p{0.32\textwidth}
    >{\raggedright\arraybackslash}p{0.40\textwidth}@{}}
    \toprule
    No. & Target object and context & Semantic question & Candidate answers \\
    \midrule
    1 & A form with credential inputs and login controls &
    What is the semantic function of the detected form element? &
    \texttt{insufficient\_\allowbreak evidence};
    \texttt{collects\_\allowbreak user\_\allowbreak credentials};
    \texttt{submits\_\allowbreak configuration\_\allowbreak data};
    \texttt{displays\_\allowbreak error\_\allowbreak messages};
    \texttt{handles\_\allowbreak media\_\allowbreak upload} \\
    \addlinespace
    2 & A submission operation linked to the form &
    What is the primary operational purpose of the detected event handler? &
    \texttt{insufficient\_\allowbreak evidence};
    \texttt{initiates\_\allowbreak page\_\allowbreak load};
    \texttt{handles\_\allowbreak user\_\allowbreak interaction};
    \texttt{manages\_\allowbreak window\_\allowbreak dimensions};
    \texttt{triggers\_\allowbreak form\_\allowbreak submission} \\
    \addlinespace
    \multicolumn{4}{@{}p{\textwidth}@{}}{\textit{Pair 1--2:} Operation $\xrightarrow{\mathtt{targets}}$ form.} \\
    \midrule
    3 & An external script element and page language cues &
    What is the observable function role of the script identified by the src attribute? &
    \texttt{insufficient\_\allowbreak evidence};
    \texttt{dependency\_\allowbreak loading};
    \texttt{localization\_\allowbreak execution};
    \texttt{framework\_\allowbreak initialization} \\
    \addlinespace
    4 & An external resource referenced by a script, with its reference context &
    What is the observable function role of the resource identified by the path attribute? &
    \texttt{insufficient\_\allowbreak evidence};
    \texttt{styling\_\allowbreak definition};
    \texttt{favicon\_\allowbreak representation};
    \texttt{application\_\allowbreak bundle};
    \texttt{localization\_\allowbreak script} \\
    \addlinespace
    \multicolumn{4}{@{}p{\textwidth}@{}}{\textit{Pair 3--4:} Script $\xrightarrow{\mathtt{references}}$ resource.} \\
    \midrule
    5 & A response text span with HTTP status and device management login text &
    What is the scope of the content returned in the HTTP response? &
    \texttt{insufficient\_\allowbreak evidence};
    \texttt{authentication\_\allowbreak challenge};
    \texttt{management\_\allowbreak interface};
    \texttt{error\_\allowbreak message};
    \texttt{redirect\_\allowbreak location} \\
    \addlinespace
    6 & An adjacent text span with vendor, product model, and account prompts &
    Classify the primary explicit subject of the indicated object's visible content. &
    \texttt{device itself};
    \texttt{device administration interface};
    \texttt{user or account}; \texttt{network service};
    \texttt{help content}; \texttt{other};
    \texttt{insufficient\_\allowbreak evidence} \\
    \addlinespace
    \multicolumn{4}{@{}p{\textwidth}@{}}{\textit{Pair 5--6:} The \texttt{next\_text\_span} relation connects adjacent spans.} \\
    \bottomrule
  \end{tabular}
  \par\medskip
  \noindent\textbf{(b) Example request and DSEJEV output}\par\smallskip
  \noindent\begin{minipage}[t]{0.55\textwidth}
    \vspace{0pt}
    \raggedright
    \textbf{Target operation:} \texttt{onchange} event.\par
    \textbf{Related control:} \texttt{select} element with \texttt{id} and \texttt{name} set to \texttt{LanguageSelect}, \texttt{class} set to \texttt{width120}, and \texttt{onchange} set to \texttt{g\_oLogin.\allowbreak changeFrameLanguage(\allowbreak this.value)}.\par
    \textbf{Related script:} \texttt{g\_oLogin.\allowbreak changeFrameLanguage(\allowbreak this.value)}.\par
    \textbf{Relations:} Control $\xrightarrow{\mathtt{targets}}$ operation; operation $\xrightarrow{\mathtt{references}}$ script.\par
    \textbf{Question:} What is the primary operational intent of the event handler?
  \end{minipage}\hfill
  \begin{minipage}[t]{0.43\textwidth}
    \vspace{0pt}
    \setlength{\tabcolsep}{0.02\linewidth}
    \begin{tabular}{@{}
      >{\raggedright\arraybackslash}p{0.12\linewidth}
      >{\raggedright\arraybackslash}p{0.59\linewidth}
      >{\raggedleft\arraybackslash}p{0.21\linewidth}@{}}
      \toprule
      Option & Candidate answer & Probability \\
      \midrule
      A & \texttt{insufficient\_\allowbreak evidence} & 0.0005 \\
      B & \texttt{navigation} & 0.0278 \\
      C & \texttt{data\_submission} & 0.0040 \\
      D & \texttt{state\_change} & 0.9626 \\
      E & \texttt{authentication} & 0.0050 \\
      F & \texttt{error\_handling} & 0.0001 \\
      \bottomrule
    \end{tabular}
  \end{minipage}
  \endgroup
\end{table}
\end{minipage}
}]

\section{Ethical Considerations}
\label{sec:ethics}

We use the publicly released DevTag and BLMProbe datasets~\cite{wan2023devtag,tian2025blmprobe} to evaluate vendor and device-type identification from recorded Web responses. The experiments involve no additional scans or interactions with live devices. We evaluate semantic questions on stored responses and analyze form and script relationships without executing the corresponding actions. We report aggregate identification and efficiency results. ValQID is intended to support authorized network asset inventory. Device identification can also aid reconnaissance, so operators should use ValQID only on devices and networks they are authorized to assess.

\section{Semantic Questions and Relation Templates}
\label{app:semantic-questions}

Table~\ref{tab:semantic-question-examples}(a) presents six questions from the final libraries, grouped by object roles and connections into three pairs. Each question is evaluated with its target object and local context. Panel~(b) shows an example request and the option probabilities returned by DSEJEV.

The control name and handler identify a language-switching operation, and DSEJEV assigns the highest probability to \texttt{state\_\allowbreak change}. ValQID attaches the full option distribution to the target object for identity prediction and subsequent computation selection. Example~2 in Appendix~\ref{app:identification-examples} describes the corresponding identification process.

\section{Representative Identification Examples}
\label{app:identification-examples}

\noindent\textbf{Example 1: Two rounds of semantic evaluation.} The response contains \texttt{zyxel vsg1432-b101}, \texttt{web configurator}, and username and password prompts. ValQID first selects the question ``What is the primary operational role of the target resource?'' with a predicted value of 0.1170; DSEJEV assigns probability 0.6323 to \texttt{web\_\allowbreak authentication\_\allowbreak gateway}.

After updating the identification state, ValQID selects ``What specific function does the target perform regarding network connectivity?'' with a predicted value of 0.1147; DSEJEV assigns probability 0.999993 to \texttt{insufficient\_\allowbreak evidence}. With both option distributions added to the identification state, ValQID combines this state with response features and label-pair information to predict \textbf{zyxel / router}, matching the reference labels.

\noindent\textbf{Example 2: A contextual semantic query.} The response contains \texttt{Server: NVR Webserver}, username and password controls, and the language-selection control and event handler shown in Table~\ref{tab:semantic-question-examples}(b). ValQID selects a question about the intent of the \texttt{onchange} handler, with a predicted value of 0.4968. Using the associated control and script, DSEJEV classifies the operation as \texttt{state\_change} with probability 0.9626.

After the option distribution updates the relational state, the highest predicted value among the remaining candidate groups is $-0.1996$, so ValQID stops querying. The final predictor combines this state with response features and label-pair information to output \textbf{Hanbanggaoke / DVR}, matching the reference labels.

\section{Data and Implementation Details}
\label{app:experimental-details}

\subsection{Data Splits and Evaluation}

\noindent\textbf{Data splits.} For DevTag, we start from the released training, development, and test partitions and group duplicate records by exact response text. Text groups spanning multiple partitions are assigned to one partition, with priority given to test, development, and training, in that order. Groups with conflicting labels are excluded. After selecting HTTP responses, we retain the vendor and device-type classes present in all three partitions to form closed-set identification tasks.

BLMProbe is partitioned by source host, with target proportions of 60\%, 20\%, and 20\% for training, development, and testing. Source groups connected by identical responses are merged and assigned to one partition under the same priority rule. The number of responses in each partition depends on the merged group sizes.

\noindent\textbf{Training data use.} ValQID further divides each training set into two disjoint subsets: one for training the relation-based identity model and discovering semantic conditions, and the other for fitting the question-group value model. These subsets contain 47,140 and 11,784 responses for DevTag, and 5,257 and 1,315 responses for BLMProbe. Responses sharing a source group or identical full text remain in the same subset. We train the value model using identification gains computed with the identity model frozen. The final joint character classifier uses all training responses with valid vendor--type labels: 58,924 for DevTag and 6,407 for BLMProbe. The DevTag nearest-neighbor component uses the 47,140 responses assigned to identity-model training.

\noindent\textbf{Evaluation.} All methods in Table~\ref{tab:main-comparison} are evaluated on the same test responses within each dataset, using a common label vocabulary and scoring procedure. We evaluate vendor, device-type, and joint identification on all 8,490 DevTag test responses. For BLMProbe, device-type evaluation uses all 2,370 test responses; vendor and joint evaluation use the 2,347 responses with valid labels for both tasks. Joint accuracy requires both labels to be correct for the same response. Vendor and device-type macro-averaged metrics are computed over the fixed class vocabulary of each dataset.

\subsection{Model Training and Inference}

\noindent\textbf{Model training.} DSEJEV is adapted from JevK5 4B using LoRA, with Qwen3.5-397B-A17B providing supervision for local semantic questions. We set the LoRA rank, scaling factor, and dropout to 16, 32, and 0.05, respectively, and train for two epochs with a learning rate of $3\times10^{-5}$. Each parameter update uses 128 questions, with a maximum input length of 2,048 tokens per request.

The relation-based identity model uses a batch size of 256, with learning rates of $5\times10^{-5}$ for DevTag and $2\times10^{-4}$ for BLMProbe. The question-group value model is trained with AdamW, a learning rate of $10^{-4}$, a batch size of 512, and Huber loss with $\delta=1$. Model checkpoints and fusion parameters are selected on the development set.

\noindent\textbf{Inference environment.} End-to-end efficiency experiments run on a single NVIDIA H100 80GB GPU. DSEJEV uses BF16 precision with the LoRA weights merged into the base model. The system processes 64 responses per batch. Each semantic batch contains at most 64 requests and 16,384 tokens after padding. Four CPU worker processes prepare the inputs.

\end{document}